\documentclass[10pt,twocolumn,letterpaper]{article}

\usepackage{cvpr}      

\definecolor{cvprblue}{rgb}{0.21,0.49,0.74}
\usepackage[pagebackref,breaklinks,colorlinks,allcolors=cvprblue]{hyperref}

\def\paperID{3045} 
\def\confName{CVPR}
\def\confYear{2025}

\title{RevSAM2: Prompt SAM2 for Medical Image Segmentation via Reverse-Propagation without Fine-tuning}

\author{Yunhao Bai\textsuperscript{1} \quad Boxiang Yun\textsuperscript{1} \quad Zeli Chen\textsuperscript{2} \quad Qinji Yu\textsuperscript{3} \quad Yingda Xia\textsuperscript{2} \quad Yan Wang\textsuperscript{1}\footnotemark[1]\\
\normalsize{\textsuperscript{1}Shanghai Key Laboratory of Multidimensional Information Processing, East China Normal University}\\
\normalsize{\textsuperscript{2}DAMO Academy, Alibaba Group}\\
\normalsize{\textsuperscript{3}Shanghai Jiao Tong University, Shanghai, China.}\\
}

\begin{document}
\maketitle
\begin{abstract}
The Segment Anything Model 2 (SAM2) has recently demonstrated exceptional performance in zero-shot prompt segmentation for natural images and videos.  
However, when the propagation mechanism of SAM2 is applied to medical images, it often results in spatial inconsistencies, leading to significantly different segmentation outcomes for very similar images.
In this paper, we introduce RevSAM2, a simple yet effective self-correction framework that enables SAM2 to achieve superior performance in unseen 3D medical image segmentation tasks without the need for fine-tuning. Specifically, to segment a 3D query volume using a limited number of support image-label pairs that define a new segmentation task, we propose reverse propagation strategy as a query information selection mechanism. Instead of simply maintaining a first-in-first-out (FIFO) queue of memories to predict query slices sequentially, reverse propagation selects high-quality query information by leveraging support images to evaluate the quality of each predicted query slice mask. The selected high-quality masks are then used as prompts to propagate across the entire query volume, thereby enhancing generalization to unseen tasks. Notably, we are the first to explore the potential of SAM2 in label-efficient medical image segmentation without fine-tuning. Compared to fine-tuning on large labeled datasets, the label-efficient scenario  provides a cost-effective alternative for medical segmentation tasks, particularly for rare diseases or when dealing with unseen classes.
Experiments on four public datasets demonstrate the superiority of RevSAM2 in scenarios with limited labels, surpassing state-of-the-arts by {12.18\%} in Dice. The code will be released.\vspace{-0.8em}
\end{abstract}    
\section{Introduction}
\label{sec:intro}

\begin{figure}
    \centering
    \includegraphics[width=\linewidth]{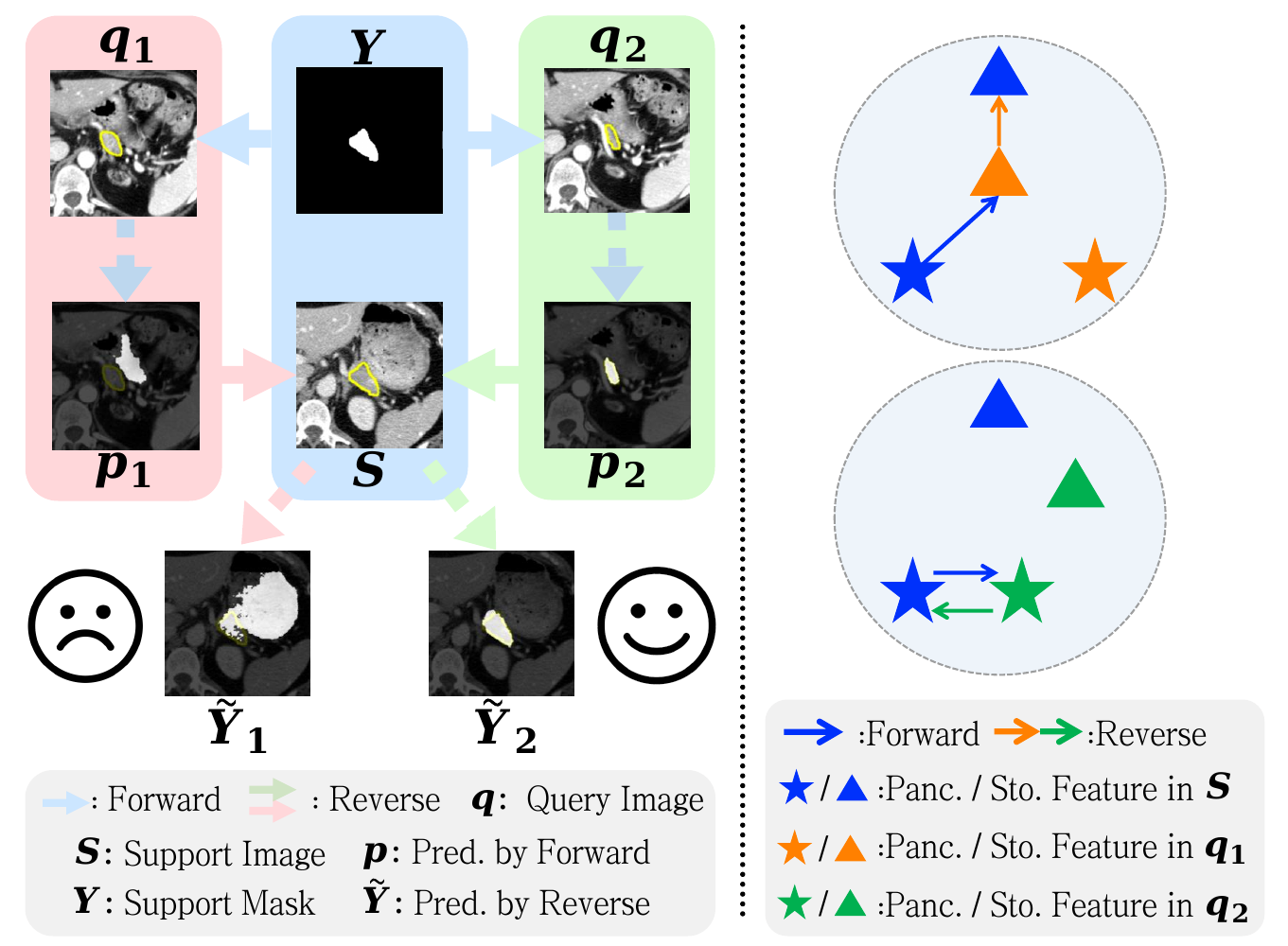}\vspace{-0.5em}
    \caption{Example of reverse propagation (left) and in feature space (right). $\textbf{\textit{S}}$ is a CT slice and $\textbf{\textit{Y}}$ is its segmentation mask, while $\textbf{\textit{q}}_1$ and $\textbf{\textit{q}}_2$ are two adjacent CT slices from a different CT scan than $\textbf{\textit{S}}$. The pancreatic tail region in all three images is outlined in yellow. The prediction masks $\textbf{\textit{p}}_1$ and $\textbf{\textit{p}}_2$ correspond to $\textbf{\textit{q}}_1$ and $\textbf{\textit{q}}_2$, respectively, generated by the memory bank that stores features of $\textbf{\textit{S}}$ and $\textbf{\textit{Y}}$. Conversely, $\widetilde{\textbf{\textit{Y}}}_1$(~$\widetilde{\textbf{\textit{Y}}}_2$~) is the prediction masks for $\textbf{\textit{S}}$ generated by the memory bank filled with the features of $\textbf{\textit{p}}_1$ and $\textbf{\textit{q}}_1$ ($\textbf{\textit{p}}_2$ and $\textbf{\textit{q}}_2$). In our framework, $\textbf{\textit{p}}_1$ will be discarded, and $\textbf{\textit{q}}_2$ along with $\textbf{\textit{p}}_2$ are used to support the re-segmentation of $\textbf{\textit{q}}_1$.}
    \vspace{-0.9em}
    \label{fig:figure1}
\end{figure}

The Segment Anything Model 2 (SAM2) \cite{ravi2024sam2} has shown remarkable zero-shot prompt segmentation capabilities in natural images and videos. With points, boxes or mask as prompts, SAM2 can accurately segment the object foreground within an image and track the object across frames in a video. However, similar to its predecessor, SAM \cite{kirillov2023segany}, SAM2 struggles with medical images, particularly in Computed Tomography (CT) and Magnetic Resonance Imaging (MRI) scans. This limitation stems from the lack of medical images in its training data, resulting in the model’s inability to precisely delineate the foreground of organs and other structures in medical images via their semantic features.

Recent studies \cite{medsam2,medicalsam2,sam2adapterevaluatingadapting} have sought to adapt SAM2 for medical imaging applications. Like fine-tuning on SAM, these approaches typically involve fine-tuning some components of SAM2, such as the mask decoder, using a certain amount of labeled data. However, these methods face two primary limitations: first, they require a substantial amount of labeled data and considerable time for training, which cannot handle unseen tasks or settings under domain shift; second, even after training, they still depend on interactive prompts to perform segmentation on the target images. Moreover, only taking cues from SAM, these methods inevitably overlook the fact that compared with SAM, SAM2 has made meaningful enhancements in its network architecture, resulting in a more capable and versatile successor. 
SAM2 uses a memory attention module to access information stored in the memory bank about the target object, enabling it to segment the corresponding target in subsequent images. This inspires us to explore it from a different perspective, \emph{i.e.}, is it possible to leverage SAM2’s ability to automatically segment targets in medical images such as CT and MRI (referred to as query images) based on only a few similar images, specifically a few 2D slices (called support images), and their labels, without \emph{any} fine-tuning? {This setting shows promising research potential in many clinical scenarios, }such as in cases where clinical researchers often lack the resources and expertise to train neural networks \cite{universeg}, only a few or scarce labels can be provided for certain rare diseases or working with unseen classes.


{Compared to training a model from scratch \cite{nnunet, swinunetr} using a few labeled slices, fine-tuning a foundation model \cite{samed,hsam,hqsam,catsam} with such data seems more feasible. However, the extremely limited amount of data still poses a significant challenge. Some few-shot methods \cite{aasdcl,srcl,rpt,GMRD} attempt to perform inference in scenarios with limited unseen class data, but these methods still require a large number of data from other classes to train a support-query segmentation model.}


In this paper, we fully leverage the memory bank and propagation ability of SAM2. Since SAM2 is trained on natural images and videos, giving it the ability to locate objects in target images even when facing certain positional and appearance changes. A straightforward approach to segment 3D medical images based on a few labeled support images using SAM2 is to store the support images and their masks in the memory bank, and then maintain a first-in-first-out (FIFO) queue of memories to predict each query slice. However, this approach may lead to inferior performance if previous query slices are not well segmented. This is because, when applied to unseen medical images, SAM2 often overemphasizes the positional and appearance information of target tissues stored in the memory bank. As a result, incorrect segmentation occurs when target tissues experience positional or appearance shifts, especially if the stored information is not generalizable. Although tissues in CT and MRI images, such as organs, tend to share similar locations and appearances among patients, considerable variations exist due to patient heterogeneity, making segmentation challenging \emph{without} fine-tuning. This observation led us to consider whether it is possible to automatically select well-segmented query slices (conditional slices) and propagate them to slices without prompts (non-conditional slices) \emph{within the same query volume}.

Without any groundtruth masks, deciding whether a query slice is well segmented seems challenging, since the IoU score predictor in SAM2 trained specifically on natural images is not reliable for evaluating medical images. To solve this problem, we propose a \textbf{surprisingly simple yet effective} reverse-propagation strategy. Concretely, we first forward propagate the features of support images and masks to obtain the prediction for each slice in the query volume. Then, we reverse-propagate the query image along with its predicted mask back to derive masks for the support images. The quality of the query slice prediction is evaluated using the Dice scores between the predicted masks and the ground-truth masks of support images. Fig.~\ref{fig:figure1} {(left)} illustrates the motivation of our simple process. {As shown in Fig.~\ref{fig:figure1}, features of $\textbf{\textit{S}}$ and $\textbf{\textit{Y}}$ propagates an incorrect prediction $\textbf{\textit{p}}_1$ for a query slice $\textbf{\textit{q}}_1$ but a correct prediction $\textbf{\textit{p}}_2$ for its adjacent query slice $\textbf{\textit{q}}_2$. After reverse propagation, $\textbf{\textit{q}}_1$ and $\textbf{\textit{p}}_1$ lead to an even more incorrect prediction $\widetilde{\textbf{\textit{Y}}}_1$ in $\textbf{\textit{S}}$, while $\textbf{\textit{q}}_2$ and $\textbf{\textit{p}}_2$ yield an accurate prediction  $\widetilde{\textbf{\textit{Y}}}_2$. 
This intriguing phenomenon can be explained in Fig~.\ref{fig:figure1} {(right)} in the feature space: for the query slice $\textbf{\textit{q}}_1$, the feature of the real target pancreas ($\color{orange}{\bigstar}$) may encounter some positional or appearance shift which deviates from the pancreas' feature of the support image stored in the memory ($\color{blue}{\bigstar}$) by a certain distance. This makes SAM2 to find a feature of another tissue stomach ($\color{orange}{\blacktriangle}$) closer to the pancreas' feature ($\color{blue}{\bigstar}$). During reverse propagation, the predicted target in the $\textbf{\textit{q}}_1$ ($\color{orange}{\blacktriangle}$) has a higher chance to find another closer tissue \emph{e.g.}, stomach ($\color{blue}{\blacktriangle}$) in the support image, rather than pancreas ($\color{blue}{\bigstar}$). But for the query slice $\textbf{\textit{q}}_2$, since the distance between the feature of the pancreas in the support image ($\color{blue}{\bigstar}$) and the one in $\textbf{\textit{q}}_2$ ($\color{green}{\bigstar}$) is small, there is a high probability that after reverse propagation, the pancreas ($\color{blue}{\bigstar}$) can still be identified.

Based on the proposed reverse-propagation, we design RevSAM2, a self-correction framework, which prompts SAM2 for medical image segmentation via reverse-propagation without fine-tuning.
{Given only a few 2D support images and their corresponding segmentation masks, our framework can segment the query volume without additional prompts or retraining, adapting quickly by simply changing the support images.}
We validate RevSAM2 on four publicly available multiorgan datasets, demonstrating its effectiveness by achieving state-of-the-art (SOTA) performance on all datasets. Furthermore, we also validated the robustness of RevSAM2 for domain adaptation in scenarios with limited labels.\vspace{-0.5em}
\section{Related Work}
\subsection{Segment Anything Model 2}
Compared to Segment Anything Model \cite{kirillov2023segany} (SAM), which focuses solely on promptable image segmentation, Segment Anything Model 2 (SAM2) introduces an additional capability for promptable video segmentation. The components responsible for image segmentation in SAM2 remain the same as those in SAM: the image encoder, prompt encoder, and mask decoder. In SAM2, the image encoder and prompt encoder independently encode the input image and prompt information, which are then fused and passed into the mask decoder to generate the segmentation mask.

For video segmentation, SAM2 incorporates additional components: memory encoder, memory bank, and memory attention. The workflow is as follows: First, \textbf{add prompt}. Prompts are applied to select frames (conditional frames), which undergo independent image segmentation to produce masks. Second, \textbf{fill memory bank}. The memory encoder encodes the features of images and masks, storing them in the memory bank. Finally, \textbf{propagate}. For non-conditional frames, image features are extracted and processed with memory attention using the features stored in memory bank. After generating the mask, step two is repeated.

\subsection{SAM-based Medical Image Methods}
Segment Anything Model \cite{kirillov2023segany} (SAM) has been extensively trained on over a billion natural images, demonstrating strong zero-shot segmentation capabilities for natural images: given a prompt (e.g., a point or a bounding box) for an image, it can segment the object foreground indicated by the prompt. Some previous works \cite{chai2023ladder,samu,inputsam} attempted to apply SAM to medical images, but experiments \cite{samonmedical,samondigital,whensammeets,sammd} have shown that SAM's zero-shot segmentation performance significantly drops on unseen medical images. Therefore, recent studies have focused on how to fine-tune SAM using a certain amount of medical images to adapt it to medical images \cite{samed,catsam,hqsam,hsam,samadapter}. For instance, MedSAM \cite{medsam} created a large-scale medical image dataset to retrain SAM with bounding box prompts; SAMed \cite{samed} introduced LoRA \cite{hu2021lora} layers into the image encoder while using the original mask decoder; H-SAM \cite{hsam} used LoRA-inserted image encoders to enhance SAM's feature extraction capability for medical images and designed a hierarchical mask decoder to enable prompt-free segmentation.

Similar to SAM, SAM2 also performs suboptimally on medical images when using point and box prompts. After SAM2 was released, several studies built upon the successful adaptations of SAM for medical imaging, aiming to adapt SAM2 for the medical image domain. For example, MedicalSAM2 \cite{medicalsam2} fine-tune the all components of SAM2 exclude prompt encoder, MedSAM \cite{medsam2} only fine-tune the image encoder and mask decoder, while SAM2-Adapter \cite{sam2adapterevaluatingadapting} introduces lightweight adapters into the image encoder, which are fine-tuned alongside the mask decoder during weight updates, {using 4*A100 GPUs for training.}

In the SAM2 architecture, the image encoder is significantly more complex than the mask decoder. The success of these approaches, which primarily focus on fine-tuning the mask decoder, suggests that SAM2's image encoder is already capable of effectively encoding the information present in medical images. In contrast to these methods, our approach aims to adapt SAM2 to medical imaging in a more challenging setting: {insufficient label and} without any weight fine-tuning.\vspace{-0.5em}

\begin{figure*}[htbp]
    \centering
    \vspace{-0.5em}
    \includegraphics[width=\linewidth]{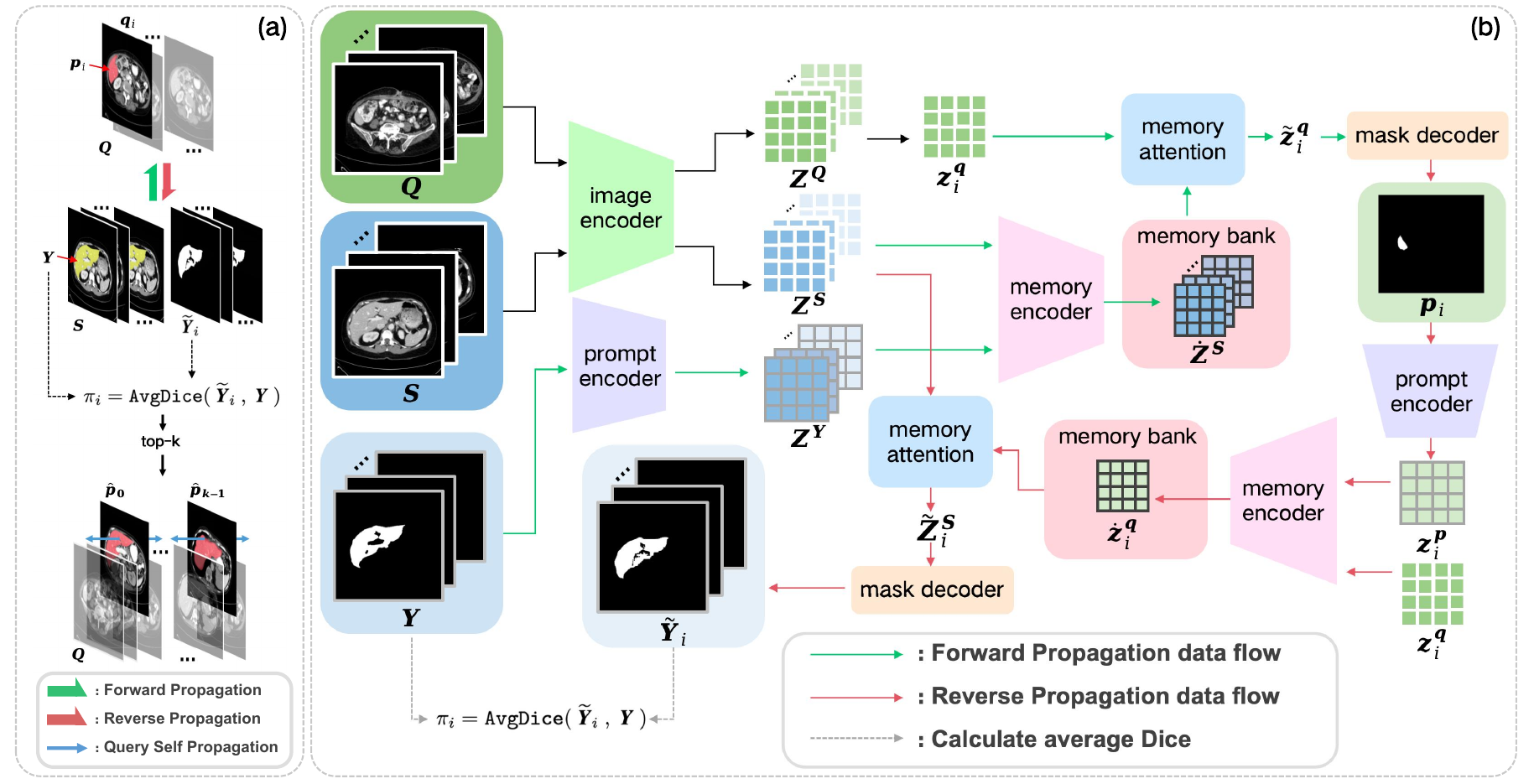}\vspace{-0.5em}
    \caption{The overall framework of RevSAM2 (a) and illustration of forward propagation and reverse propagation (b). To evaluate the quality of the prediction $\textbf{\textit{p}}_i$ obtained by forward propagating $\textbf{\textit{S}}$ and $\textbf{\textit{Y}}$ onto $\textbf{\textit{q}}_i$, we reverse propagate $\textbf{\textit{q}}_i$ and $\textbf{\textit{p}}_i$ back to $\textbf{\textit{S}}$ to obtain $\widetilde{\textbf{\textit{Y}}}_i$, and calculate the average dice $\pi_i$ between $\widetilde{\textbf{\textit{Y}}}_i$ and $\textbf{\textit{Y}}$ and treat it as the as the metric to evaluate the accuracy of $\textbf{\textit{p}}_i$.}
    \vspace{-0.6em}
    \label{fig:framwork}
\end{figure*}


\subsection{Label Insufficient Medical Image Segmentation}
Label Insufficient Medical Image Segmentation is an emerging field that addresses the challenge of limited annotated data—a common issue in the medical domain due to the high costs, complexity of annotation, and legal constraints on data sharing. Specifically, a major research focus within this field is Few-Shot Medical Image Segmentation (FSMIS). FSMIS techniques can be broadly categorized into two main types: prototypical network-based models \cite{wang2019panet,ssl_alpnet,adnet,qnet,yu2021location} and two-branch interaction-based models \cite{senet,mrrnet,GCN-DE,CRAPNet}. These methods train a model to segment the query image by drawing information from the support image and its mask, and then validate the model on unseen categories that are not part of the training process.


However, most of these methods treat consecutive slices as independent entities, segmenting each slice separately in a few-shot manner. We believe that in a sequence of consecutive slices, the segmentation results of earlier slices can be leveraged to assist in the segmentation of subsequent slices.\vspace{-0.5em}
\section{Method}

Mathematically, we first define a 3D volume as $\textbf{\textit{X}}$, and its $i$th slice is denoted as $\textbf{\textit{x}}_i$. Given the 3D volume of a query medical image as $\textbf{\textit{Q}} \in \mathbb{R}^{M \times W \times H}$, a set of few 2D support images as $\textbf{\textit{S}} \in \mathbb{R}^{N \times W \times H}$ in the 3D form, and their corresponding segmentation labels as $\textbf{\textit{Y}} \in \{0, 1\}^{N \times W \times H}$, which indicate the background and target regions in $\textbf{\textit{S}}$. Our goal is to predict the per-voxel segmentation map $\widetilde{\textbf{\textit{P}}} \in \mathbb{R}^{M \times W \times H}$ of $\textbf{\textit{Q}}$ using the few support image-label pairs. Our segmentation framework, RevSAM2, is built upon SAM2.

The overall pipeline of the proposed RevSAM2 is illustrated in Fig. \ref{fig:framwork} (a). For the $i$th slice $\textbf{\textit{q}}_i$ in volume \textbf{\textit{Q}}, we first obtain its prediction $\textbf{\textit{p}}_i$ by using the memory bank which stores the features of  $\textbf{\textit{S}}$ and $\textbf{\textit{Y}}$ through memory attention mechanism. We then assess the quality of $\textbf{\textit{p}}_i$ using reverse propagation, which treats $\textbf{\textit{q}}_i$-$\textbf{\textit{p}}_i$ pair as a support image-label pair, and predict segmentation map $\widetilde{\textbf{\textit{Y}}}_i$ for all support 2D images $\textbf{\textit{S}}$. After obtaining the average predicted Dice scores for $\textbf{\textit{S}}$, we select the top $k$ images with the highest scores. 
These retained images, along with their corresponding predictions, serve as conditional images to propagate information to all slices within \textbf{\textit{Q}} through self-propagation, ultimately resulting in prediction $\widetilde{\textbf{\textit{P}}}$.

\subsection{Forward Propagation}
\label{sec:forw prop}

As illustrated in Fig. \ref{fig:framwork} (b), we first use an image encoder $\mathcal{E}^\mathtt{I}$, parameterized by $\mathbf{\Theta}^\mathtt{I}$, and a prompt encoder $\mathcal{E}^\mathtt{P}$, parameterized by $\mathbf{\Theta}^\mathtt{P}$ to encode $\textbf{\textit{S}}$ and $\textbf{\textit{Y}}$, respectively, to obtain image features $\textbf{\textit{Z}}^\textbf{\textit{S}}$ and prompt features $\textbf{\textit{Z}}^\textbf{\textit{Y}}$:
\begin{align}
&\textbf{\textit{Z}}^\textbf{\textit{S}}=\mathcal{E}^\mathtt{I}(~\textbf{\textit{S}}~,~\mathbf{\Theta}^\mathtt{I}~),\label{eq:S image feature}\\
&\textbf{\textit{Z}}^\textbf{\textit{Y}}=\mathcal{E}^\mathtt{P}(~\textbf{\textit{Y}}~,~\mathbf{\Theta}^\mathtt{P}~).
\end{align}
After feature extracting of support images and their segmentation labels, memory encoder sums $\textbf{\textit{Z}}^\textbf{\textit{Y}}$ element-wise with $\textbf{\textit{Z}}^\textbf{\textit{S}}$, followed by a lightweight convolutional layer, whose parameters are $\mathbf{\Theta}^\mathtt{M}$, to fuse the information and obtain $\dot{\textbf{\textit{Z}}}^\textbf{\textit{S}}$:
\begin{equation}
\dot{\textbf{\textit{{Z}}}}^\textbf{\textit{S}}=\mathcal{E}^\mathtt{M}(~\textbf{\textit{Z}}^\textbf{\textit{S}}~,~ \textbf{\textit{Z}}^\textbf{\textit{Y}}~,~ \mathbf{\Theta}^\mathtt{M}~),
\end{equation}
where $\mathcal{E}^\mathtt{M}(\cdot,\cdot,\mathbf{\Theta}^\mathtt{M})$ means the memory encoder.

The memory bank retains information about the segmentation of $\textbf{\textit{S}}$ for the target object segmentation in $\textbf{\textit{Q}}$. Specifically, for the $i$th slice $\textbf{\textit{q}}_i$ in $\textbf{\textit{Q}}$ ($0 \leq i < M$), its image feature embedding $\textbf{\textit{z}}^\textbf{\textit{q}}_i$ is extracted by the image encoder as described for $\textbf{\textit{S}}$ in Eq. \ref{eq:S image feature}: $\textbf{\textit{z}}^\textbf{\textit{q}}_i=\mathcal{E}^\mathtt{I}(~\textbf{\textit{q}}_i~,~\mathbf{\Theta}^\mathtt{I}~).$
Then, conditioning $\textbf{\textit{z}}^\textbf{\textit{q}}_i$ on $\dot{\textbf{\textit{{Z}}}}^\textbf{\textit{S}}$ through self-attention and cross-attention in the memory attention mechanism, we obtain the fused vision feature $\widetilde{\textbf{\textit{z}}}^\textbf{\textit{q}}_i$, which is formulated as:
\begin{equation}
    \widetilde{\textbf{\textit{z}}}^\textbf{\textit{q}}_i=\mathcal{A}(~\textbf{\textit{z}}^\textbf{\textit{q}}_i~|~\dot{\textbf{\textit{{Z}}}}^\textbf{\textit{S}},~\mathbf{\Omega}~),
\end{equation}
where $\mathcal{A}(\cdot | \cdot, \mathbf{\Omega})$ means the memory attention module whose parameters are $\mathbf{\Omega}$.
Finally, $\widetilde{\textbf{\textit{z}}}^\textbf{\textit{q}}_i$ is fed into the mask decoder $\mathcal{D}^\mathtt{M}$ to obtain the predicted segmentation map $\textbf{\textit{p}}_i$:
\begin{equation}
    \textbf{\textit{p}}_i=\mathcal{D}^\mathtt{M}(~\widetilde{\textbf{\textit{z}}}^\textbf{\textit{q}}_i~, ~\mathbf{\Gamma}^\mathtt{M}~),
\end{equation}
where $\mathbf{\Gamma}^\mathtt{M}$ indicate the parameter of the mask decoder.

Thus, we obtain the segmentation mask $\textbf{\textit{p}}_i$ of $\textbf{\textit{q}}_i$ based on the information from $\textbf{\textit{S}}$ and $\textbf{\textit{Y}}$, without any direct prompt from $\textbf{\textit{q}}_i$. However, due to the spatial positional perturbation of SAM2 mentioned in Sec. \ref{sec:intro}, the quality of $\textbf{\textit{p}}_i$ may not be optimal. Next, we propose a novel reverse-propagate strategy to evaluate the quality of $\textbf{\textit{p}}_i$ by checking whether it can reverse propagate back to $\textbf{\textit{S}}$ and generate $\textbf{\textit{Y}}$.

\subsection{Reverse Propagation}
\label{sec:rev prop}

After obtaining $\textbf{\textit{p}}_i$, we apply reverse propagation to calculate the predicted Dice scores for $\textbf{\textit{S}}$ using $\textbf{\textit{q}}_i$-$\textbf{\textit{p}}_i$ pair as a support image-label pair. This simple yet effective evaluation strategy can determine whether $\textbf{\textit{p}}_i$ is accurate or affected by positional perturbation. Similar to how we obtain $\dot{\textbf{\textit{{Z}}}}^\textbf{\textit{S}}$ from $\textbf{\textit{S}}$ and $\textbf{\textit{Y}}$, we first derive $\dot{\textbf{\textit{z}}}^\textbf{\textit{q}}_i$ from $\textbf{\textit{q}}_i$ and $\textbf{\textit{p}}_i$ as follows:\vspace{-1em}

\begin{align} 
    \textbf{\textit{z}}^\textbf{\textit{p}}_i &= \mathcal{E}^\mathtt{P}(~\textbf{\textit{p}}_i~,~\mathbf{\Theta}^\mathtt{P}~), \\
    \dot{\textbf{\textit{z}}}^\textbf{\textit{q}}_i &= \mathcal{E}^\mathtt{M}(~\textbf{\textit{z}}^\textbf{\textit{q}}_i~, ~\textbf{\textit{z}}^\textbf{\textit{p}}_i~,~\mathbf{\Theta}^\mathtt{M}~). 
\end{align}

The resulting $\dot{\textbf{\textit{z}}}^\textbf{\textit{q}}_i$ is then used to reversely segment $\textbf{\textit{S}}$. Similarly, the image features $\textbf{\textit{Z}}^\textbf{\textit{S}}$ of $\textbf{\textit{S}}$ are processed through the memory attention module, combined with $\dot{\textbf{\textit{z}}}^\textbf{\textit{q}}_i$, to generate the fused vision features $\widetilde{\textbf{\textit{Z}}}^\textbf{\textit{S}}_i$:

\vspace{-0.5em}
\begin{equation} 
    \widetilde{\textbf{\textit{Z}}}^\textbf{\textit{S}}_i = \mathcal{A}(~\textbf{\textit{Z}}^\textbf{\textit{S}}~|~\dot{\textbf{\textit{z}}}^\textbf{\textit{q}}_i~,~\mathbf{\Omega}~). 
\end{equation}

Note that the $i$th $\textbf{\textit{q}}_i$-$\textbf{\textit{p}}_i$ pair yields $\widetilde{\textbf{\textit{Z}}}^\textbf{\textit{S}}_i$. 
The prediction map $\widetilde{\textbf{\textit{Y}}}_i$ is then obtained by feeding $\widetilde{\textbf{\textit{Z}}}^\textbf{\textit{S}}_i$ to the mask decoder:

\begin{equation} 
    \widetilde{\textbf{\textit{Y}}}_i = \mathcal{D}^\mathtt{M}(~\widetilde{\textbf{\textit{Z}}}^\textbf{\textit{S}}_i~,~\mathbf{\Gamma}^\mathtt{M}~). 
\end{equation}

{The purpose of reverse propagation is to amplify the influence of wrong direction of propagation. If $\textbf{\textit{p}}_i$ is a correct segmentation, it is likely to support the accurate segmentation of $\textbf{\textit{S}}$. Conversely, an incorrect segmentation of $\textbf{\textit{q}}_i$ may lead to significant errors in the segmentation of $\textbf{\textit{S}}$, resulting in predictions that deviate substantially from $\textbf{\textit{Y}}$.} 
Thus, we calculate the average dice score between $\widetilde{\textbf{\textit{Y}}}_i$ and $\textbf{\textit{Y}}$, as a metric  to evaluate the accuracy of $\textbf{\textit{p}}_i$:

\begin{equation} 
    \pi_i =  \mathtt{AvgDice}(~\widetilde{\textbf{\textit{Y}}}_i~,~\textbf{\textit{Y}}~).
\end{equation}
After obtaining all scores of $\textbf{\textit{P}}$, we retain the top $k$ scoring masks among all $\textbf{\textit{p}}_i$ in $\textbf{\textit{P}}$ to get the final volume prediction map for $\textbf{\textit{Q}}$ by query self propagation (described in Sec.~\ref{sec:queryself})

{Intuitively, when there are more support images, the feature bias of each support image become smaller, making the response Dice obtained through reverse propagation more representative and accurate. This aspect will be further discussed in the ablation study.}

\begin{figure}
    \centering
    \includegraphics[width=0.9\linewidth]{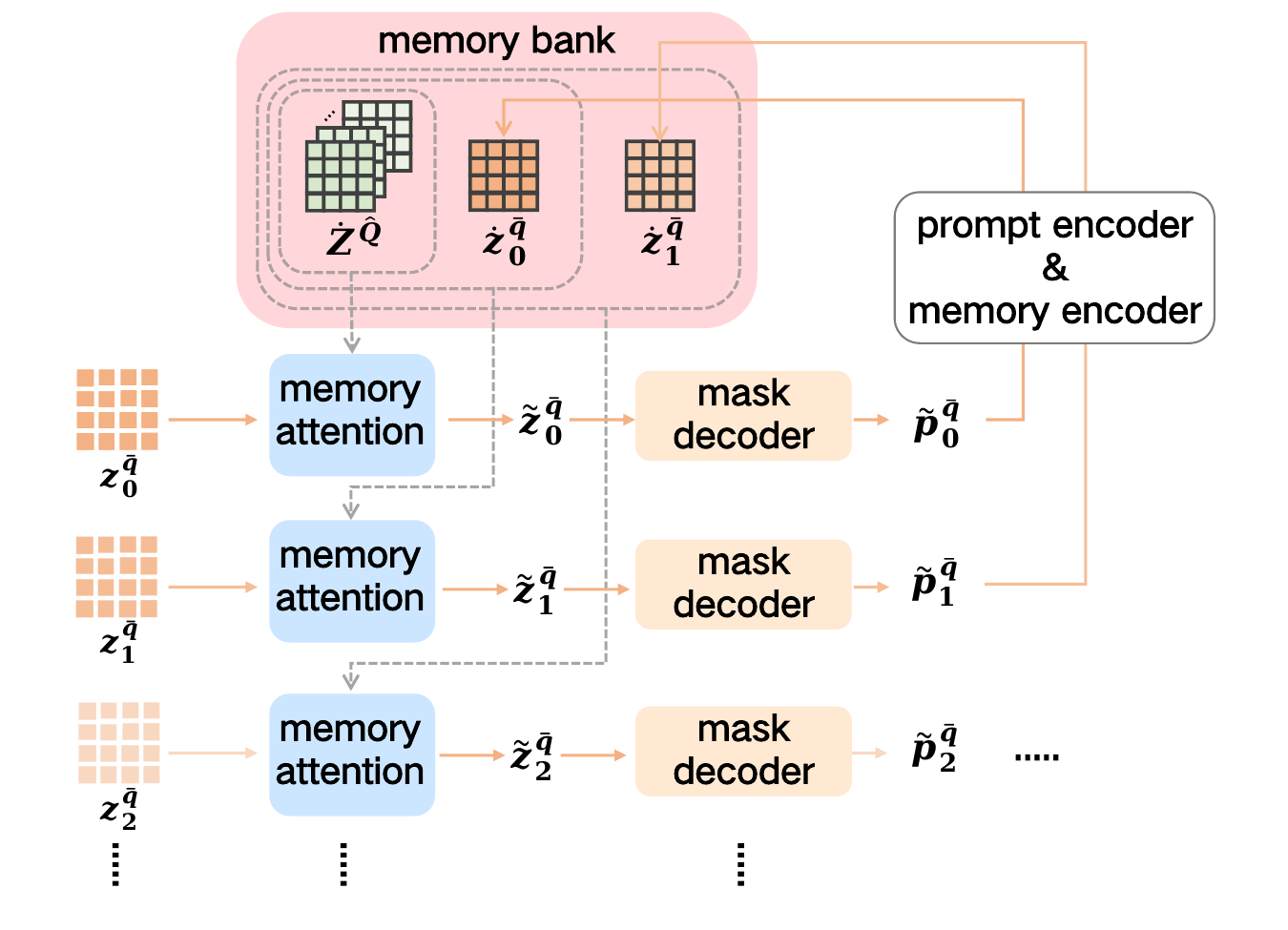}\vspace{-1em}
    \caption{The illustration of the query self propagation. In query self propagation, the memory bank continuously stores the features of conditional slices selected by reverse propagation, while maintaining a FIFO queue to store the features of non-conditional slices during internal query inference.}\vspace{-1.2em}
    \label{fig:queryself_propagate}
\end{figure}

\subsection{Query Self Propagation}
\label{sec:queryself}
The overall framework of query self propagation is illustrated in Fig. \ref{fig:queryself_propagate}. We refer to the query slices with the top $k$ scoring prediction masks as conditional query slices, denoted as $\hat{\textbf{\textit{Q}}}$, where $\hat{\textbf{\textit{Q}}} \in \mathbb{R}^{k \times W \times H}$, and their corresponding prediction masks as $\hat{\textbf{\textit{P}}}$. We set $k=7$ as default .The remaining query slices are referred to as {n}on-{c}onditional query slices, whose 3D volume is denoted as $\bar{\textbf{\textit{Q}}}$.

We first extract the image features $\textbf{\textit{Z}}^{\hat{\textbf{\textit{Q}}}}$ 
and prompt features $\textbf{\textit{Z}}^{\hat{\textbf{\textit{P}}}}$, and then combine them to obtain the feature $\dot{\textbf{\textit{{Z}}}}^{\hat{\textbf{\textit{Q}}}}$ stored in the memory bank, similar to $\dot{\textbf{\textit{Z}}}^\textbf{\textit{S}}$ in propagation and $\dot{\textbf{\textit{z}}}^\textbf{\textit{q}}_i$ in reverse propagation:\vspace{-0.5em}
\begin{align}
    \textbf{\textit{Z}}^{\hat{\textbf{\textit{Q}}}} &= \mathcal{E}^\mathtt{I} (~\hat{\textbf{\textit{Q}}}~,~\mathbf{\Theta}^\mathtt{I}~), \\
    \textbf{\textit{Z}}^{\hat{\textbf{\textit{P}}}} &= \mathcal{E}^\mathtt{P} (~\hat{\textbf{\textit{P}}}~,~\mathbf{\Theta}^\mathtt{P}~), \\
    \dot{\textbf{\textit{{Z}}}}^{\hat{\textbf{\textit{Q}}}} &= \mathcal{E}^\mathtt{M} (~\textbf{\textit{Z}}^{\hat{\textbf{\textit{Q}}}}~,~ \textbf{\textit{Z}}^{\hat{\textbf{\textit{P}}}}~,~\mathbf{\Theta}^\mathtt{M}~).
\end{align}

\vspace{-0.3em}
Unlike the static memory bank used in the forward propagation and reverse propagation, the memory bank in query self propagation not only retains information about the conditional query slices but also maintains a first-in-first-out (FIFO) queue of memories of the $\tau$ most recent non-conditional query slices and their predicted masks. Here, we follow the default configuration $\tau=7$ of SAM2 TINY Model. {For each non-conditional query slice $\bar{\textbf{\textit{q}}}_i$, its image feature $\textbf{\textit{z}}^{\bar{\textbf{\textit{q}}}}_i$ is extracted by image encoder:} $\textbf{\textit{z}}^{\bar{\textbf{\textit{q}}}}_i = \mathcal{E}^\mathtt{I} (~\bar{\textbf{\textit{q}}}_i~,~\mathbf{\Theta}^\mathtt{I}~).$
Then, we obtain the fused vision feature of the $j$th non-conditional query slice $\bar{\textbf{\textit{q}}}_j$ via:
\begin{equation}
    \widetilde{\textbf{\textit{z}}}^{\bar{\textbf{\textit{q}}}}_j =
    \mathcal{A}(~\textbf{\textit{z}}^{\bar{\textbf{\textit{q}}}}_j~|~\dot{\textbf{\textit{{Z}}}}^{\hat{\textbf{\textit{Q}}}}~,~\dot{\textbf{\textit{{z}}}}^{\bar{\textbf{\textit{q}}}}_{j-1}, \ldots, \dot{\textbf{\textit{{z}}}}^{\bar{\textbf{\textit{q}}}}_{\max(j-\tau,0)},~\mathbf{\Omega}~).
\end{equation}
{We use recent non-conditional query slices to reduce the feature bias, as the positional and appearance shifts between conditional and non-conditional slices are smaller than those between support images and non-conditional slices.} Finally, we obtain the prediction mask $\widetilde{\textbf{\textit{p}}}^{\bar{\textbf{\textit{q}}}}_j$ for $\bar{\textbf{\textit{q}}}_j$:
\begin{equation}
    \widetilde{\textbf{\textit{p}}}^{\bar{\textbf{\textit{q}}}}_j = \mathcal{D}^\mathtt{M}(~\widetilde{\textbf{\textit{z}}}^{\bar{\textbf{\textit{q}}}}_j~,~\mathbf{\Gamma}^\mathtt{M}~).
\end{equation}\vspace{-1.5em}

All non-conditional prediction masks $\widetilde{\textbf{\textit{P}}}^{\bar{\textbf{\textit{Q}}}}$ are combined with the conditional masks $\hat{\textbf{\textit{P}}}$ to form the final volume prediction mask $\widetilde{\textbf{\textit{P}}}$ for the query $\textbf{\textit{Q}}$.

\section{Experiments}

\begin{table*}[t]
\renewcommand\arraystretch{0.9}
\centering
\resizebox{\linewidth}{!}
{
\begin{tabular}{c|cccccccccccccc|cc}
\toprule
Method             & spleen         & kidnetR        & kidneyL        & gall    & eso      & liver          & stomach        & arota          & IVC       & veins           & pancreas       & AG R           & AG L           & duode       & mDSC    &mNSD       \\ \midrule
nnU-Net \cite{nnunet} (ALL)      & 96.14             & 88.19             & 81.15             & 84.86             & 78.14             & 90.41             & 85.05             & 91.32             & 87.92             & 70.17             & 77.87             & 73.32             & 75.15             & 77.95             & 82.70        & 84.40     \\
Swin UNETR \cite{swinunetr} (ALL)   & 95.69             & 87.80             & 95.31             & 80.60             & 80.37             & 89.39             & 76.66             & 90.21             & 86.85             & 66.24             & 74.71             & 72.68             & 75.80             & 71.30             & 81.69           &83.27   \\ \midrule
1pos 1neg points   & 56.55          & 60.39          & 63.07          & 16.44          & 6.40           & 60.90          & 45.14          & 50.46          & 17.58          & 10.71          & 10.57          & 2.35           & 1.21           & 12.26          & 29.57         & 13.53 \\
3 pos 3 neg points & 62.73          & 73.87          & 73.98          & 15.77         & 12.52          & 68.82          & 47.38          & 48.34          & 17.06          & 7.72           & 11.39          & 1.75           & 1.82           & 11.47          & 32.47         & 15.40  \\
bbox               & 86.41          &  81.91         & 80.56          & 64.04          & 34.18          & 76.86          & 47.89          & 39.47          & 38.24          & 33.62          & 31.31          & 25.10          & 22.77          & 36.54          & 49.92        & 40.56  \\
mask               & 91.67          & 87.72          & 81.37          & 66.83          & 36.36          & 79.71          & 50.39          & 87.45          & 69.27          & 47.72          & 28.84          & 26.47          & 38.81          & 45.54          & 59.87         & 56.16 \\ \midrule
nnU-Net \cite{nnunet}            & 18.00             & 25.40             & 30.60             & 7.66             & 10.27             & 32.49             & 12.33             & 35.85             & 14.06             & 11.31             & 18.46            & 4.75             & 11.91             & 8.47             & 17.27           & 16.18  \\
Swin UNETR \cite{swinunetr}         & 20.90             & 19.92             & 8.40             & 6.32             & 14.79             & 33.56             & 11.81             & 28.38             & 16.15             & 8.23             & 15.92             & 7.31             & 21.95             & 11.45             & 16.08           & 13.58  \\
UniverSeg \cite{universeg}          &58.95           & 65.25          &64.40           & 24.43          & 22.82          & 77.10          & 41.61          & 47.03          & 35.86          & 27.62          & 21.75          & 15.56          & 12.27          & 17.78        & 38.03        & 45.34  \\
SAMed \cite{samed}              & 71.59          & 47.11          & 52.64          & 31.83          & 32.36          & 84.03          & 40.30          & 65.99          & 36.39          & 37.24          & 24.96          & 4.99          & 8.04          & 15.42          & 39.49        & 33.85  \\
H-SAM \cite{hsam}              & {\ul 87.93}    & {\ul 74.74}          & {\ul 81.95}          & 55.75          & 45.42          & {\ul 89.70}          & 55.05          & {\ul 80.09}          & 52.81          & 42.68          & 28.80          & 29.97          & 31.72          & 26.51          & 55.94       & 50.60   \\
HQ-SAM \cite{hqsam}             & 74.54          & 69.54          & 73.97          & {\ul 60.35}          & \textbf{74.05}           & 67.41          & 63.38          & 62.97          & {\ul 75.46}          & 14.14          & 46.97         & {\ul 36.01}              &   \textbf{51.80}             & 40.97          & 57.97         &  48.41 \\
CAT-SAM \cite{catsam}            & 72.53          & 52.26       & 60.73         & 58.97          & 51.13          & 87.33          & \textbf{75.08}     & 63.62    & 65.06      & \textbf{61.52}         & \textbf{59.93}         & 30.86     & 41.57   & \textbf{57.08}     & {\ul 59.83}   &  {\ul 55.46} \\
MedicalSAM2 \cite{medicalsam2}           &   78.66         & 73.90       & 69.85         & 56.13          & 45.08          & 85.49          & 61.63     & 30.18    & 50.49      & 27.32         & 45.75         & 33.03     & 23.42   & {\ul 46.40}     & 51.95   &  38.61 \\
RevSAM2(ours)   & \textbf{93.99} & \textbf{82.50}    & \textbf{85.78}    & \textbf{81.19} & {\ul 55.42}    & \textbf{92.85} & {\ul 70.70} & \textbf{86.07} & \textbf{78.21} & {\ul 54.91} & {\ul 57.96}    & \textbf{50.29} & {\ul 45.20} & 42.97    & \textbf{69.86}  & \textbf{67.66}  \\ \bottomrule
\end{tabular}
}
\vspace{-0.5em}
\caption{Comparison of mDSC (\%) and mNSD (\%) on the BTCV dataset. `ALL' refers to using the full training data to demonstrate the upper bound. For each category, only 10 slices are used as the training set or support images, with all methods using the same slices for comparison. The best performances are highlighted in \textbf{bold}, while the second-best performances are indicated with \underline{underlines}.}
\label{tab:BTCV dataset}\vspace{-0.8em}
\end{table*}

\subsection{Datasets and evaluation}

In order to validate our method, we conduct experiments on each single-organ segmentation task across four multi-organ medical image segmentation datasets, including BTCV \cite{btcv}, AbdomenCT-1K \cite{AbdomenCT-1K}, Synapse-CT \cite{synapseCT} and CHAOS-MRI \cite{chaos}. 

\noindent\textbf{BTCV} ~ The Multi-Atlas Labeling Beyond the Cranial Vault (BTCV) challenge dataset contains 50 abdominal CT scans. Initially, the dataset includes annotations for 13 organs, with additional annotations for the duodenum added in \cite{btcvlabel}. 
For our study, we use the same data split and 14 multi-organ labels as in \cite{clip-driven}.

\noindent\textbf{AbdomenCT-1K} ~ This dataset consists of 1112 CT scans from five different sources, with annotations for the liver, kidney, spleen, and pancreas. \cite{abdomenctlabel} extended the label set to 13 organs. Like BTCV, we adopt the same data split and 13 multi-organ labels as used in \cite{clip-driven}.

\noindent\textbf{Synapse-CT} ~ Synapse-CT is an abdominal CT dataset acquired from the MICCAI 2015 Multi-Atlas Abdomen Labeling Challenge, which comprises 30 3D abdominal CT scans. Following GMRD \cite{GMRD}, we evaluate RevSAM2 on left kidney, right kidney, liver, and spleen. 

\noindent\textbf{CHAOS-MRI} ~ CHAOS-MRI is an abdominal MRI dataset obtained from the ISBI 2019 Combined Healthy Abdominal Organ Segmentation Challenge. It contains 20 3D T2-SPIR MRI scans. Following GMRD \cite{GMRD}, we also evaluate RevSAM2 on the left kidney, right kidney, liver and spleen.

We conduct individual segmentation experiments for each class in the datasets. Unless otherwise specified, we randomly select 10 separated slices from three volumes in the training set for each class to simulate a real-world label insufficient scenario. 
To mitigate the effect of random selection, we use three groups of random slices, and report the average performance. The details of each group are provided in the supplementary material. 
The final evaluation metrics for our method are mean Dice Similarity Coefficient (mDSC) and mean Normalized Surface Dice (mNSD). 

\subsection{Comparison Methods}
Our approach employs the SAM2 TINY model, which is the most lightweight variant. Initially, we test its prompt segmentation capabilities on medical images using points (1 positive and 1 negative, 3 positive and 3 negative), bounding box (bbox), and mask. For point prompts, we randomly select pixels from the ground truth as positive points and chose negative points from the false positives generated by the segmentation of positive points, repeating this process for each slice. Regarding bounding boxes and masks, we adhere to SAM2's evaluation method: generating bounding boxes based on the ground truth of the first slice or using the ground truth directly as mask prompts, allowing automatic propagation to obtain volume predictions.

We then conduct comparative experiments with four different approaches. The \textbf{first} type of approach involves training from scratch, such as nnU-Net \cite{nnunet} and Swin UNETR \cite{swinunetr}. As expected, with only 10 labeled slices, these methods struggled to train models with good generalization. We also train them using all available training data to demonstrate their upper bounds. The \textbf{second} approach is a universal few-shot segmentation model, UniverSeg\cite{universeg}, which has been trained on a large number of medical image datasets to create a support-query style segmentation model. However, due to its limitation to 128x128 resolution input images, we have to compress the images for comparison. The \textbf{third} type of approach indicate SAM-based fine-tuning, including SAMed \cite{samed}, H-SAM \cite{hsam}, HQ-SAM \cite{hqsam}, CAT-SAM \cite{catsam}, and MedicalSAM2 \cite{medicalsam2}. These methods leverage foundation models (SAM \cite{kirillov2023segany} or SAM2 \cite{ravi2024sam2}) that have been extensively trained on natural images, making it more likely to fine-tune a segmentation model with some level of generalization in extremely data-scarce scenarios (10 slices). Notably, only SAMed and H-SAM can perform fully automatic segmentation, while other SAM-based methods still require prompts for query slices. \textbf{Finally}, we compare our results with traditional few-shot methods, including AAS\_DCL \cite{aasdcl}, SR\&CL \cite{srcl}, RPT \cite{rpt} and GMRD \cite{GMRD}, which train on some categories within the training set and test on unseen categories in the test set. It is worth noting that for these comparisons, we strictly follow their one-shot strategy, employing their exact support-query selection method. More details are illustrated in supplementary material.

\begin{table*}[t]
\renewcommand\arraystretch{0.9}
\centering
\resizebox{\linewidth}{!}
{
\begin{tabular}{c|ccccccccccccc|cc}
\toprule
Method             & liver          & kidneyR        & spleen         & pancreas       & arota          & IVC       & AGL            & AGR            & gall           & eso            & stomach        & duode       & kidneyL        & mDSC     &mNSD     \\ \midrule
nnU-Net \cite{nnunet} (ALL)      & 99.03             & 98.29             & 99.01             & 90.46             & 97.65             & 95.16             & 90.64             & 89.79             & 93.05            & 90.11             & 97.38             & 91.85            & 98.17             & 94.70        & 95.70     \\
Swin UNETR \cite{swinunetr} (ALL)   & 98.95             & 98.18             & 98.92             & 88.26             & 97.45             & 95.38             & 90.45             & 89.35             & 92.22             & 88.47             & 96.49             & 88.78             & 98.30             & 94.00          & 94.50   \\ \midrule
1pos 1neg points   & 77.25          & 85.89          & 79.18          & 17.19          & 51.02          & 12.24          & 1.00           & 4.18           & 15.25          & 6.06           & 35.47          & 20.06          & 78.79          & 37.20       & 18.93   \\
3 pos 3 neg points & 79.54          & 95.37          & 90.62          & 12.45          & 57.83          & 20.14          & 1.72           & 2.36           & 12.34          & 11.04          & 43.58          & 15.13          & 89.30          & 40.88       & 24.34   \\
bbox               & 89.60          & 82.48          & 92.01          & 53.56          & 33.72          & 54.16          & 25.75          & 26.83          & 45.75          & 22.25          & 64.07          & 41.52          & 84.94          & 55.13       & 37.48   \\
mask               & 90.00          & 95.30          & 95.62          & 36.05          & 91.42          & 79.19          & 27.41          & 54.39          & 62.06          & 70.55          & 62.81          & 34.67          & 94.01          & 68.73       & 61.42   \\ \midrule
nnU-Net \cite{nnunet}            & 71.52             & 76.22             & 46.93             & 24.36             & 61.35             & 57.44             & 21.21             & 31.95             & 12.90             & 25.73             & 11.82             & 11.53             & 68.00             & 40.64          & 29.78   \\
Swin UNETR \cite{swinunetr}         & 53.84             & 43.18             & 53.44             & 21.88             & 41.58             & 31.68             & 28.29             & 15.62             & 14.02             & 23.03             & 12.22             & 5.94             & 48.48             & 30.25          & 23.35   \\
UniverSeg \cite{universeg}          & 89.51          & 77.86          & 77.87          & 29.29          & 58.60          & 50.16          & 23.49          & 29.01          & 38.98          & 44.78          & 56.81          & 33.57          & 77.03         & 52.84        & 58.48   \\
SAMed \cite{samed}              & 89.51         & 74.36          & 69.81          & 38.03          & 68.86          & 35.51          & 35.71          & 11.37          & 60.50          & 56.67          & 48.66          & 25.49           & 79.30          & 53.37       & 39.20   \\
H-SAM \cite{hsam}              & 90.32         & {\ul 93.76}          & {\ul 91.15}           & 51.25          & {\ul 79.41}          & 69.98          & {\ul 54.76}          & 42.43          & 71.07          & 59.16          & 74.30          & 44.06          & {\ul 92.00}          & {\ul 70.28}        & {\ul 58.68}  \\
HQ-SAM \cite{hqsam}             & 71.17          & 68.96          & 71.04          & 39.54          & 69.10          & \textbf{84.76}          & 42.65          & 8.86          & 67.54           & {\ul 67.91}          & 70.31         & 46.71              & 80.23              & 60.68       &  42.55   \\
CAT-SAM \cite{catsam}            & 84.91          & 82.17        & 80.37         & {\ul 66.76}         & 67.58         & 65.14         & 41.96         & {\ul 47.05}          & 67.81          & 59.62         & {\ul 81.18} & \textbf{61.22}    & 80.93     & 68.21   &  56.28   \\
MedicalSAM2 \cite{medicalsam2}            & {\ul 95.18}          & 76.09        & 90.57         & 59.60         & 29.20         & 54.96         & 43.28         & 40.49          & {\ul 79.35}          & 62.16         & 74.68 & {\ul 59.14}    & 79.38     & 64.93   &  44.06   \\
RevSAM2(ours)   & \textbf{95.49} & \textbf{96.23}          & \textbf{97.07}    & \textbf{68.92} & \textbf{93.16}    & {\ul 81.78} & \textbf{67.65} & \textbf{71.53} & \textbf{83.80} & \textbf{75.32} & \textbf{90.74}    & 54.85 & \textbf{95.48} & \textbf{82.46}       & \textbf{77.33}    \\ \bottomrule
\end{tabular}
}
\vspace{-0.5em}
\caption{Comparison of mDSC (\%) and mNSD (\%) on the AbdomenCT-1K dataset with other methods, only 10 slices are used as the training set or support images.}
\label{tab:abdomenct dataset}\vspace{-0.8em}
\end{table*}

\begin{table}[t]
\renewcommand\arraystretch{0.9}
\centering
\resizebox{0.83\linewidth}{!}
{
\begin{tabular}{c|cc|cc}
\toprule
Query Dataset    & \multicolumn{2}{c|}{BTCV}              & \multicolumn{2}{c}{AbdomenCT}          \\ \midrule
Support Dataset  & \multicolumn{1}{c|}{BTCV}  & Abd. & \multicolumn{1}{c|}{Abd.} & BTCV  \\ \midrule
UniverSeg\cite{universeg}        & \multicolumn{1}{c|}{38.83} & 44.53 & \multicolumn{1}{c|}{52.84}     & 52.49 \\
SAMed \cite{samed}            & \multicolumn{1}{c|}{39.67} & 40.74 & \multicolumn{1}{c|}{53.37}     & 44.76    \\ 
H-SAM \cite{hsam}            & \multicolumn{1}{c|}{56.96} & 55.04 & \multicolumn{1}{c|}{{\ul 70.28}}     & 63.04    \\ 
HQ-SAM \cite{hqsam}           & \multicolumn{1}{c|}{{\ul 61.34}} & {\ul 59.52} & \multicolumn{1}{c|}{60.68}     & 59.39    \\ 
CAT-SAM \cite{catsam}          & \multicolumn{1}{c|}{59.70} & 59.50 & \multicolumn{1}{c|}{68.21}     & {\ul 68.10}    \\ 
MedicalSAM2 \cite{medicalsam2}          & \multicolumn{1}{c|}{53.85} & 54.10 & \multicolumn{1}{c|}{64.93}     & 65.26    \\
Ours & \multicolumn{1}{c|}{\textbf{71.01}} & \textbf{69.03} & \multicolumn{1}{c|}{\textbf{82.46}}     & \textbf{77.47} \\ \bottomrule
\end{tabular}
}\vspace{-0.5em}
\caption{Domain adaptation mDSC (\%) comparison results on the BTCV and AbdomenCT datasets. Query Dataset: the dataset of test images; Support Dataset: the dataset of support images.}
\label{tab:domain adaptation}\vspace{-1.5em}
\end{table}

\subsection{Main Results}

\noindent\textbf{BTCV and AbdomenCT} ~ Table \ref{tab:BTCV dataset} and Table \ref{tab:abdomenct dataset} show the results on BTCV dataset and AbdomenCT dataset, respectively. RevSAM2 shows outstanding segmentation ability with insufficiency labels (10 slices for each organ) on both datasets. As shown in the tables, both the re-trained models (nnUNet, Swin UNETR) and the SAM-based fine-tuning methods fail to achieve satisfactory segmentation performance, even for those using bounding box prompts (HQ-SAM, CAT-SAM, and MedicalSAM2) per slice. This is due to the extreme scarcity of labeled data, preventing these methods from converging to a model with good generalization. For UniverSeg, as it can only handle low-resolution images of 128x128, it performs well on larger organs (\emph{e.g.}, liver: 77.10\% on BTCV, 89.51\% on AbdomenCT) but poorly on smaller organs (\emph{e.g.}, right adrenal gland: 15.56\% on BTCV, 29.01\% on AbdomenCT), leading to unsatisfactory overall segmentation performance. Notably, compared to other methods, SAM2, without any fine-tuning, achieves comparable results by utilizing mask prompts and the memory attention mechanism, demonstrating the strength of SAM2's memory mechanism and laying a solid foundation for the success of our training-free method. Compared to other methods, RevSAM2 achieves significant improvements in mDSC on the BTCV and AbdomenCT datasets, with gains of 10.03\% (69.86\% vs. 59.83\%) and 12.18\% (82.46\% vs. 70.28\%), respectively.

\begin{table}[t]
\renewcommand\arraystretch{0.9}
\resizebox{\linewidth}{!}
{
\begin{tabular}{c|c|ccccc}
\toprule
Method   & Dataset                     & spleen & liver & LK    & RK    & mean  \\ \midrule
AAS-DCL \cite{aasdcl}  & \multirow{5}{*}{Synapse-CT} & 72.30  & 78.04 & 74.58 & 73.19 & 74.52 \\
SR\&CL \cite{srcl}   &                             & 73.41  & 76.06 & 73.45 & 71.22 & 73.53 \\
RPT \cite{rpt}      &                             & {\ul 79.13}  & {\ul 82.57} & 77.05 & 72.58 & 77.83 \\
GMRD \cite{GMRD}     &                             & 78.31  & 79.60 & {\ul 81.70} & {\ul 74.46} & {\ul 78.52} \\
Ours     &                             & \textbf{94.02}  & \textbf{89.41} & \textbf{84.20} & \textbf{82.05} & \textbf{87.42} \\ \midrule
AAS\_DCL \cite{aasdcl} & \multirow{5}{*}{CHAOS-MRI}  & 76.24  & 72.33 & 80.37 & 86.11 & 78.76 \\
SR\&CL \cite{srcl}   &                             & 76.01  & 80.23 & 79.34 & 87.42 & 80.77 \\
RPT \cite{rpt}      &                             & {\ul 76.37}  & {\ul 82.86} & 80.72 & {\ul 89.82} & 82.44 \\
GMRD \cite{GMRD}     &                             & 76.09  & 81.42 & \textbf{83.96} & \textbf{90.12} & {\ul 82.90} \\
Ours     &                             & \textbf{85.82}  & \textbf{88.44} & {\ul 81.63} & 85.22 & \textbf{85.28} \\ \bottomrule
\end{tabular}
}\vspace{-0.5em}
\caption{Comparison of mDSC (\%) with other few-shot methods on Synapse-CT (top) and CHAOS-MRI (bottom). We test the one-shot setting of these methods, strictly following their support-query pair selection strategies.}\vspace{-0.5em}
\label{tab:sabs and chaos}
\end{table}

\noindent\textbf{Domain Adaptation}
The BTCV dataset includes annotations for 14 organs, which encompass all 13 organs labeled in the AbdomenCT dataset. 
We further demonstrate the strong robustness of our method when using support images from different datasets. In Table \ref{tab:domain adaptation}, we conduct the following experiments: for segmenting 13 organs in BTCV (excluding veins), we use images from AbdomenCT as support images; similarly, for segmenting 13 organs in AbdomenCT, we use images from BTCV as support images. We compare these results with the domain adaptation capabilities of other methods. We use the same three groups of support images as in Tables \ref{tab:BTCV dataset}-\ref{tab:abdomenct dataset}, and the average results are presented. As shown in Table~\ref{tab:domain adaptation}, RevSAM2 still achieves the best performance in the domain adaptation experiments compared to other methods (69.03\% vs. 59.52\% of HQ-SAM on BTCV, 77.47\% vs. 68.10\% of CAT-SAM on AbdomenCT). Notably, the strong generalization ability of UniverSeg allows it to achieve better segmentation performance on BTCV when using AbdomenCT as support, compared to directly using BTCV. This result may indicate that AbdomenCT provides a broader or more consistent representation that enhances the model's ability to generalize across datasets. However, the overall segmentation performance remains somewhat unsatisfactory, potentially due to the loss of information caused by image compression.

\noindent\textbf{Synapse-CT and CHAOS-MRI} ~ As shown in the Table \ref{tab:sabs and chaos}, RevSAM2 achieves state-of-the-art mDSC performance on both Synapse-CT and CHAOS-MRI datasets, with improvements of 8.9\% (87.42\% vs. 78.52\%) and 2.38\% (85.28\% vs. 82.90\%). Especially for the spleen, it achieves improvements of 14.89\% (94.02\% vs. 79.13\%) and 9.45\% (85.82\% vs. 76.37\%) on Synapse-CT and CHAOS-MRI, respectively. It is worth mentioning that the support and query images for compared methods are paired one-to-one. For fair comparison, we stack all query images corresponding to the same support image to implement our method. More details are provided in the supplementary material.

\begin{table}[t]
\renewcommand\arraystretch{0.9}
\centering
\resizebox{0.9\linewidth}{!}
{
\begin{tabular}{cccc|ccc}
\toprule
\multirow{2}{*}{\begin{tabular}[c]{@{}c@{}}Forw \\ Prop\end{tabular}} & \multirow{2}{*}{\begin{tabular}[c]{@{}c@{}}Query \\ Info\end{tabular}} & \multirow{2}{*}{\begin{tabular}[c]{@{}c@{}}Rand\\ Select\end{tabular}} & \multirow{2}{*}{\begin{tabular}[c]{@{}c@{}}Rev\\ Prop\end{tabular}} & \multicolumn{3}{c}{Support image number} \\ \cline{5-7} 
                      &                                                                        &                                                                        &                                                                     & 10        & 5        & 1        \\ \midrule
\checkmark                    &                                                                        &                                                                        &                                                                     & 64.93     & 61.00    & 40.62    \\
\checkmark                    & \checkmark                                                                     &                                                                        &                                                                     & 64.68     & 58.69        &  38.51       \\
\checkmark                    & \checkmark                                                                      & \checkmark                                                                      &                                                                     &  65.31         &  61.45        &    40.26      \\
\checkmark                     & \checkmark                                                                      &                                                                        & \checkmark                                                                   & 69.86     & 67.38    & 51.57    \\ \bottomrule
\end{tabular}
}\vspace{-0.5em}
\caption{Ablation study on the BTCV dataset. Forw Prop: Forward Propagation; Query Info: Using adjacent query slice information; Rand Select: Randomly selecting query slices as the conditional slices; Rev Prop: Selecting the conditional slices via reverse prop.}
\label{tab:abaltion}\vspace{-0.5em}
\end{table}

\begin{figure}[t]
    \centering
    \includegraphics[width=\linewidth]{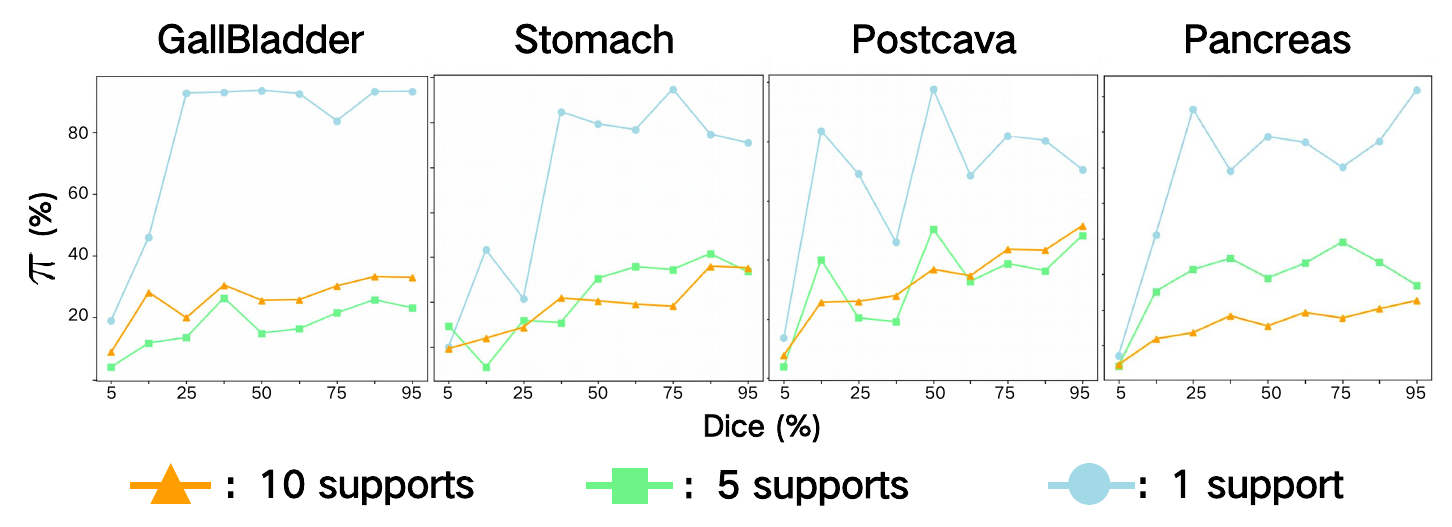}\vspace{-0.5em}
    \caption{Line charts of $\pi$ (\%) versus the actual Dice (\%) of $\textbf{\textit{p}}$ on the BTCV dataset when the number of supports is 10, 5, and 1.}\vspace{-0.8em}
    \label{fig:responce-dice}
\end{figure}

\subsection{Ablation Study}
To verify the effectiveness of the two key steps in our method: (1) utilizing query volume's own information during inference (Query Self Propagation); and (2) selecting high quality query prediction via reverse propagation, we conduct the following ablation experiments, results are shown in Table \ref{tab:abaltion}.

\noindent\textbf{Baseline} ~  Only the forward propagation stage in Sec \ref{sec:forw prop} is used, and the dice score is calculated directly after obtaining $\textbf{\textit{P}}$. Similar to most few-shot methods, this approach only uses the information from support images and ignores the latent information in the query volume.

\noindent\textbf{Forward Prop. with Query Information} ~  In this experiment, support images and query volume are treated as conditional and non-conditional slices in Sec \ref{sec:queryself}, respectively. The memory bank maintains a FIFO queue of query information to segment subsequent query slices, without distinguishing the quality of previous query prediction—whether good or bad. As table shows, this approach performs even worse than the baseline, as the FIFO queue gets contaminated with lower-quality predicted masks, which affects the segmentation of subsequent query images.

\noindent\textbf{Random Selection of Query Information}~In this experiment, $k$ query predictions are randomly selected to replace the top-$k$ query predictions chosen through reverse propagation in Sec. \ref{sec:rev prop}. As shown in the table, the results are nearly identical to the baseline, indicating that random selection does not effectively choose high-quality query segmentation results.

\noindent\textbf{Different Support Image Numbers} ~ As shown in the table, we conducted ablation study on support image numbers of 10, 5, and 1, and our method consistently achieved significant improvements in all cases. Figure \ref{fig:responce-dice} presents line charts showing the relationship between $\pi$ values (explained in detail in Sec \ref{sec:rev prop}) and the actual Dice scores of $\textbf{\textit{p}}$ for four organs in BTCV. It can be observed that with 10 and 5 support images, the lines show a certain degree of positive correlation, but with 1 support image, the lines fluctuate significantly, and high $\pi$ values appear even at low Dice scores, indicating that the top-$k$ selections are more likely to include lower-quality $\textbf{\textit{p}}$. This is because with only one support image in the memory bank, the memory attention mechanism is more susceptible to biases caused by the position and appearance of that support image. In other words, the information in the memory bank is less generalizable. 

{
\noindent\textbf{Number of $k$} ~ As described in Sec \ref{sec:queryself}, we set the default value $k=7$. In Table \ref{tab:k_ablation}, we conduct ablation experiments with $k=9$, $k=3$ and $k=1$ under different numbers of support images ($N$) to investigate the effect of $k$. As table shows, when $N=10$ or $N=5$, the differences between $k=7$ and $k=9$ is not sensitive, and as k decreases (at $k=3$ or $k=1$), the performance gradually declines.
This observation aligns with the phenomenon observed in Fig \ref{fig:responce-dice}: when $N=10$ or $N=5$, selecting high-quality $\textbf{\textit{p}}$ based on $\pi$ is positively correlated, more conditional query slices helps improve query volume prediction. In contrast, when $N=1$, the $\textbf{\textit{p}}$ selected based on high $\pi$ value may be of lower quality, and using more conditional query slices may hinder obtaining a good query volume prediction.
}



\begin{table}[t]
\renewcommand\arraystretch{0.9}
\centering
\resizebox{0.7\linewidth}{!}
{
\begin{tabular}{l|llll}
\toprule
   & $k=9$ & $k=7$ & $k=3$ & $k=1$ \\ \midrule
$N=10$ &69.03    & 69.86    & 68.12    & 65.36    \\
$N=5$  &66.83    & 67.38   & 66.33    & 64.82    \\
$N=1$  &50.58    & 51.57    & 52.50    & 52.01   \\ \bottomrule
\end{tabular}
}\vspace{-0.5em}
\caption{Ablation of $k$ on BTCV dataset. We conduct this ablation experiment under different support image numbers ($N$).}\vspace{-1em}
\label{tab:k_ablation}
\end{table}

\section{Conclusion}
We propose RevSAM2, enabling SAM2 to perform medical image segmentation in data-scarce scenarios without any fine-tuning. By proposing reverse propagation, our RevSAM2 is able to select high-quality query predictions and employ these predictions as mask prompts to propagate within the query. Our RevSAM2 opens a new direction for leveraging SAM2. Notably, without any fine-tuning, RevSAM2 outperforms state-of-the-art few-shot algorithms under the one-shot setting, achieving superior segmentation performance. Its potential for application in data-scarce scenarios could provide an economical and efficient solution for medical image segmentation.
\clearpage

\newpage
{
    \small
    \bibliographystyle{ieeenat_fullname}
    \bibliography{main}
}

\onecolumn

\end{document}